# Chance-Constrained Control with Lexicographic Deep Reinforcement Learning

Alessandro Giuseppi, *Member, IEEE*, and Antonio Pietrabissa, *Member, IEEE*

*Abstract*—This paper proposes a lexicographic Deep Reinforcement Learning (DeepRL)-based approach to chance-constrained Markov Decision Processes, in which the controller seeks to ensure that the probability of satisfying the constraint is above a given threshold. Standard DeepRL approaches require i) the constraints to be included as additional weighted terms in the cost function, in a multi-objective fashion, and ii) the tuning of the introduced weights during the training phase of the Deep Neural Network (DNN) according to the probability thresholds. The proposed approach, instead, requires to separately train one constraint-free DNN and one DNN associated to each constraint and then, at each time-step, to select which DNN to use depending on the system observed state. The presented solution does not require any hyper-parameter tuning besides the standard DNN ones, even if the probability thresholds changes. A lexicographic version of the well-known DeepRL algorithm DQN is also proposed and validated via simulations.

*Index Terms*—Markov decision processes, deep reinforcement learning, constrained control.

## I. Introduction and Related Work

DEEP Reinforcement Learning (DeepRL) is a branch of model-free control that is gathering great interest from the scientific community and funding institutions, thanks to the exponential increase of computing capacity availability and its capability of addressing heavily nonlinear problems starting from the analysis of the input-output pairs of a system. This paper proposes a DeepRL solution for chance-constrained control, a scenario in which the evolution of the system is steered in such a way that its constraints are satisfied with at least a certain probability threshold [1]. By imposing chance constraints, the operation of the controlled system can be confined within a certain region (e.g., for safety reasons [2]), while still allowing the state to evolve outside of that region if incentivized by an adequate economic/performance return [3], [4], or to assure the feasibility of the control.

The modelling framework utilised in this work is the one of Markov Decision Processes (MDPs), commonly used for stochastic optimization problems involving random events and decision makers [5]. The classic scenario for which MDPs were introduced is related to the solution of unconstrained optimal control problems by means of Dynamic Programming (DP) [6], but MDPs found great application in RL-based controllers [7], able to infer the optimal control law directly from experience without requiring the explicit knowledge of the system dynamics.

One of the most impactful modern contributions to DeepRL is presented in [8], [9], in which the authors demonstrated how a so-called Convolutional Deep Q-Network (DQN) was able to surpass human experts in playing a series of videogames. In the following years, DeepRL solutions found application in a broad range of domains typical of classic control systems, and were further refined with techniques inspired by classic RL theory such as Double Q-Learning [10] and actor-critic methods [11].

Several MDP studies dealt with constrained scenarios, typically by means of traditional Linear Programming and Lagrangian, or multi-objective, approaches [12], [13]. The Lagrangian approach can be used in RL/DeepRL algorithms and consists in designing a multi-objective cost function, where the constraints are translated into costs and included as additional objectives multiplied by constant weights (Lagrange multipliers). However, in model-free approaches, it is not possible to find the Lagrange multipliers by means of optimization tools: from the DeepRL viewpoint, the weights are additional hyper-parameters that have to be tuned by trial-and-error or other rather time-consuming procedures during the training phase (see [14] and references therein).

An alternative solution, at the basis of the present work, is the so-called "lexicographic" approach, already introduced in DP and RL formulations in [15], [16]. As described in Section I, in the lexicographic paradigm the action of the controller is aimed at minimizing either the primary cost function, if the system state is such that all the constraints are met, or one of the cost functions associated to the unsatisfied constraints, ordered by their relevance.

The main contribution of this paper consists in the extension of the lexicographic approach to the DeepRL domain, allowing the offline design of DeepRL-based controllers for chance-constrained systems. As detailed in Section II, besides the training of one DNN associated to the primary cost function, as in standard DeepRL, each constraint cost function constitutes the objective of an additional DNN. Even if more DNNs need

This paper was partially funded by the European Commission in the framework of the H2020 EU-Korea project 5G-ALLSTAR under Grant Agreement no. 815323

A. Giuseppi and A. Pietrabissa are both first authors and are with the Department of Computer, Control, and Management Engineering Antonio Ruberti, University of Rome La Sapienza, via Ariosto 25, 00185, Rome, Italy (email: {giuseppi, pietrabissa}@diag.uniroma1.it).

to be trained, the advantage with respect to multi-objective approaches is twofold: i) the training phase of the DNN is much simpler since there are no additional hyper-parameters (associated to the weights) to tune; ii) if the probability thresholds of the chance constraints change, the proposed algorithm can seamlessly reuse the already trained DNNs, whereas the multi-objective approaches require a new training. As detailed in Section II, besides the use of DNNs to approximate the action-value functions, the fact that they are trained offline is another difference with respect to [16], where the action-value functions are approximated online by RL algorithms.

The proposed methodology considers the class of DeepRL algorithms with discrete action space. Within this class, the methodology is independent from the chosen DeepRL algorithm and, for the sake of simplicity, is presented in Section II in a formulation based on DQN. In Section III, the approach is evaluated in an environment built from the classic cart-pole balancing problem with additional chance constraints. Section IV draws the conclusion and future works.

## I. Preliminaries on Lexicographic RL

A constrained MDP with multiple constraints is defined by the tuple $\{S, A_0, \mathbf{T}, \rho_0, \boldsymbol{\rho}, \gamma, \mathbf{K}, \mathcal{X}\}$, where: $S$ is the finite state space; $A_0$ is the finite action space (the subscript 0 is added for notation convenience); $\mathbf{T}(u) \in S \times S, \forall u \in A_0$ is the action-dependent transition probability matrix; $\rho_0: S \times A_0 \times S \to \mathbb{R}_+$ is the one-step non-negative primary cost function; $\boldsymbol{\rho}$ is a vector of one-step non-negative cost functions $\rho_c: S \times A_0 \times S \to \mathbb{R}_+$ accounting for the constraints $c = 1, \dots, C$; $\gamma$ is the discount factor, weighting immediate versus delayed costs; $\mathbf{K}$ is a vector of $C$ constant thresholds $K_c$, $c = 1, \dots, C$, each one representing the maximum tolerated expected value of the corresponding cost, as detailed afterwards; $\mathcal{X} \in X$ is the probability distribution of the initial state $s_0$ over the state set $S$ and $X$ is the set of feasible initial probability distributions. We considered deterministic policies, which associate a unique action $u \in A_0$ to each state $s \in S$. The selected action $u$ in state $s$ will be denoted as $\pi(s) = u$.

The control objective is to drive the evolution of the discrete-time Markov process $\{s_t\}_{t=1,2,\dots}$, where $s_t \in S$ is the state visited at time $t$, in order to minimize the expected discounted total cost, referred to as primary cost,

$$J^{\pi, \mathcal{X}} = E_{\mathcal{X}}\{V_0^{\pi}(s)\} = \sum_{s \in S} \chi(s) V_0^{\pi}(s), \tag{1}$$

where the operator $E_{\mathcal{X}}\{\cdot\}$ denotes the expected value under initial state distribution $\mathcal{X}$ and $V_0^{\pi}(s)$ is the state-value function in state $s$, i.e., the expected discounted total cost, with one-step cost $\rho_0$, when the initial state is $s$ and the system runs under policy $\pi$. $V_0^{\pi}(s)$ is defined as

$$V_0^{\pi}(s) \coloneqq E_{\pi}\{\sum_{t=0}^{\infty} \gamma^t \rho_0(s_t, u_t, s_{t+1}) | s_0 = s\}, \tag{2}$$

where the operator $E_{\pi}\{\cdot\}$ is the expected value when the system operates under policy $\pi$.

In constrained MDPs, additional cost functions are defined to enforce the constraints. The one-step constraint costs $\rho_c$ are used in the expected discounted total costs

$$J_c^{\pi, \mathcal{X}} = E_{\mathcal{X}}\{V_c^{\pi}(s)\} = \sum_{s \in S} \chi(s) V_c^{\pi}(s), c = 1, \dots, C, \tag{3}$$

hereafter referred to as constraint costs, with lower-bounded state-value functions, defined as

$$V_c^{\pi}(s) \coloneqq E_{\pi}\{\sum_{t=0}^{\infty} \gamma^t \rho_c(s_t, u_t, s_{t+1}) | s_0 = s\}. \tag{4}$$

Chance constraints usually limit the expected *undiscounted* constraint cost below a given threshold (e.g., in the cart-pole balancing problem of Section III, we are interested in limiting the probability that the pole angle exceeds a given threshold, regardless of when the constraint violations occur). Let $\overline{K}_c$ be the $c$-th threshold; considering that $J_c^{\pi, \mathcal{X}}$ approximates the total undiscounted expected cost scaled by $1/(1 - \gamma)$ [17][1], chance constraints can be expressed as

$$J_c^{\pi, \mathcal{X}} \leq K_c, c = 1, \dots, C, \tag{5}$$

with $K_c = \overline{K}_c / (1 - \gamma)$. The constrained MDP, with the constraints representing the chance constraints, is then formulated as the following optimization problem:

$$\begin{aligned} &\min_{\pi} J_0^{\pi, \mathcal{X}} \\ &\text{s.t. } J_c^{\pi, \mathcal{X}} \leq K_c, c = 1, \dots, C. \end{aligned} \tag{6}$$

As shown in [16], the problem (6) can be written as

$$\begin{aligned} &\min_{\pi} \sum_{s \in S} \chi(s) Q_0^{\pi}(s, \pi(s)) \\ &\text{s.t. } \sum_{s \in S} \chi(s) Q_c^{\pi}(s, \pi(s)) \leq K_c, c = 1, \dots, C, \end{aligned} \tag{7}$$

where $Q_v^{\pi}(s, u), v = 0, \dots, C$, is the state-action value function, i.e., the expected total discounted cost, with one-step cost $\rho_v$, when the initial state is $s \in S$, the initial action is $u \in A_0$ and the system runs under policy $\pi$:

$$Q_v^{\pi}(s, u) \coloneqq E_{\pi}\{\sum_{t=0}^{\infty} \gamma^t \rho_v(s_t, u_t, s_{t+1}) | (s_0, u_0) = (s, u)\} \tag{8}$$

As shown in [15], [16], the constraints are enforced by defining the vectorial action-value function

$$\boldsymbol{Q}^{\pi}(s, u) \coloneqq \begin{pmatrix} \max(K_C, Q_C^{\pi}(s, u)) \\ \vdots \\ \max(K_1, Q_1^{\pi}(s, u)) \\ Q_0^{\pi}(s, u) \end{pmatrix}. \tag{9}$$

where, without loss of generality, we assume that the constraints are ordered in ascending order of priority, i.e., the $c$-th constraint has priority over the $(c + 1)$-th one.

Under the lexicographic approach, the comparison between two policies $\pi'$ and $\pi''$ is done according to the vectorial value function (9), which, for the $c$-th element, $c = 1, \dots, C$, returns the threshold value $K_c$ if the constraint is met, the value of the corresponding state-action value function otherwise. In a generic state $s \in S$, there are three cases to consider for establishing if the policy $\pi'(s)$ is better than $\pi''(s)$, i.e.,

---

[1] This property derives from the series $\sum_{t=0}^{\infty} a^k = 1/(1-a), a \in (0,1)$.

$\pi'(s) \succ \pi''(s)$:
- if more constraints are met by $\pi'(s)$ w.r.t. $\pi''(s)$;
- if the same number $v < C$ of constraints are met by both policies and $Q_{v+1}^{\pi'}(s, \pi'(s)) < Q_{v+1}^{\pi''}(s, \pi''(s))$;
- if all the $C$ constraints are met by both policies and $Q_0^{\pi'}(s, \pi'(s)) < Q_0^{\pi''}(s, \pi''(s))$.

The overall policy $\pi'$ is better than $\pi''$ if $\pi'(s) \succeq \pi''(s)$, for all states $s \in S$, with $\pi'(s) \succ \pi''(s)$ for at least one state.

The lexicographic approach is conservative: since it checks the constraints for each possible initial state, it actually solves the following problem:

$$\min_{\pi} \sum_{s \in S} \chi(s) Q_0^{\pi}(s, \pi(s))$$
$$\text{s.t. } Q_c^{\pi}(s, \pi(s)) \leq K_c, c = 1, \ldots, C, \forall s \in S \quad (10)$$

Solving (10) leads to a conservative sub-optimal solution of problem (6). Relying on standard results on the convergence of DP and RL algorithms, the following property holds.

*Property 1* [15], [16]. By using the lexicographic approach with DP/RL algorithms, a stationary deterministic policy is found, which is lexicographically optimal with respect to the vectorial state-action value function (9).

## II. LEXICOGRAPHIC DEEP RL

### A. Training and application of L-DeepRL algorithms

In the actor-critic paradigm, a DNN (*critic*) is used to estimate the optimal state-action value function based on the observed states and costs and another DNN (*actor*) is used to estimate the optimal control action based on the observed state. In this paper, we consider the class of DeepRL algorithms implementing DNNs for the critic role only, suitable for problems with a finite action space.

In L-DeepRL algorithms, $C + 1$ critic networks are needed: one for estimating the primary value function $Q_0$ and one for each of the value functions $Q_c$'s associated to the $C$ constraints. The $(C + 1)$ DNNs are hereafter denoted as $Q_c, c = 0, \ldots, C$. For all the (finite number of) actions $u \in A_0$, the $c$-th (state,action)-value function is evaluated as $Q_c(\varphi_c(s), u|\boldsymbol{\theta}_c)$, where $\boldsymbol{\theta}_c$ is the vector collecting the parameters of the DNN and $\varphi_c(s)$ is a feature map which takes the state observations as inputs and returns the features in the feature set $\Phi_c$, i.e., $\varphi_c: S \to \Phi_c$.

The training phase is performed offline, separately for each critic network, and results in the determination of the $Q_c$'s, each one estimating the optimal state-action value function generated by the corresponding cost. In general, each DNNs could be trained according to a different algorithm. We note that, conversely, the lexicographic RL approach in [16] performs the approximation online: when the system is in a given state at time $k$, the action is chosen according to the current values of the value functions; after the observation of the cost and of the next state, the value functions are updated according to the selected RL algorithm and the new values are used for the action selection at step $k + 1$.

The key difference between DeepRL and L-DeepRL lies in the action selection strategy as, at each time-step, the controller, or RL agent, uses one of the $C + 1$ DNNs according to the lexicographic approach.

Preliminarily, for a given a policy $\pi$, the constrained action sets $A_v(s) \subseteq A_0, v = 0, \ldots, C$, are introduced:

$$A_v(s) = \{u \in A_0 | Q_c^{\pi}(\varphi_c(s), u|\boldsymbol{\theta}_c) \leq K_c, c = 1, \ldots, v\}. \quad (11)$$

By definition (11), the set $A_v(s)$ is then the set of the actions which, according to the estimated values of the action-value functions, meet the first $v$ constraints in state $s$ under policy $\pi$. If $v = 0$, the definition coincides with that of $A_0$.

The determination of the constraint action sets is straightforward, since a discrete action set $A_0$ is considered. The constraints $Q_c(\varphi_c(s), u|\boldsymbol{\theta}_c) \leq K_c, c = 1, \ldots, C$, can be verified, for the observed state $s$ and for all the actions, by simple enumeration, and the action sets $A_c(s)$ are then found by applying the definition (11). Fig. 1 reports the pseudo-code of the function, named *Function 1*, for the computation of the discrete constraint action sets.

The comparison between two policies $\pi'$ and $\pi''$ in state $s$ is done according to the lexicographic approach. Let $0 \leq v' \leq C$ be the number of ordered constraints which are met by $\pi'$ in the observed state $s$, i.e., $v'$ is such that

$$\begin{cases} Q_c^{\pi'}(\varphi_c(s), \pi'(s)|\boldsymbol{\theta}_c) \leq K_c, c = 1, \ldots, v' \\ Q_{v'+1}^{\pi'}(\varphi_{v'+1}(s), \pi'(s)|\boldsymbol{\theta}_{v'+1}) > K_{v'+1} \end{cases}, \quad (12)$$

and let $v''$ be defined accordingly for $\pi''$. Then, $\pi'(s) \succ \pi''(s)$ in the observed state $s$ if one of the following cases holds: i) $v' > v''$; ii) $v' = v'' = v < C$ and $Q_{v+1}^{\pi'} < Q_{v+1}^{\pi''}$; iii) $v' = v'' = C$ and $Q_0^{\pi'} < Q_0^{\pi''}$.

At time $t$, let the system be in state $s$. For all the actions $u \in A_0$, the L-DeepRL algorithm considers the constraints

$$Q_c^{\pi}(\varphi_c(s), u|\boldsymbol{\theta}_c) \leq K_c, c = 1, \ldots, C, \quad (13)$$

to decide whether the action selection rule of the RL algorithm must be applied considering the primary value function $Q_0$ or to one of the $C$ constraint value functions $Q_c$'s. Specifically, given the constraint action sets $A_c$'s and the number of met constraints $v$, the lexicographic action selection rule is

$$u = \begin{cases} \min_{u' \in A_C(s)} Q_0(\varphi_0(s), u'|\boldsymbol{\theta}_0) & \text{if } v = C \\ \min_{u' \in A_v(s)} Q_{v+1}(\varphi_{v+1}(s), u'|\boldsymbol{\theta}_{v+1}) & \text{otherwise} \end{cases}. \quad (14)$$

The following logic is pursued:
- if $A_C(s) \neq \emptyset$ (i.e., at least one action exists such that all the $C$ constraints are met) the controller selects an action belonging to the set $A_C(s)$ based on $Q_0(\varphi_0(s), u|\boldsymbol{\theta}_0)$ and is thus aimed at minimizing the primary cost $J_0$;
- if $A_v(s) \neq \emptyset$ and $A_{v+1}(s) = \emptyset, v = 0, \ldots, C - 1$ (i.e., at least one action exists such that the first $v$ constraints are met but no actions exist such that the first $v + 1$ constraints are met) the controller selects an action in the set $A_v(s)$ based on $Q_{v+1}(\varphi_{v+1}(s), u|\boldsymbol{\theta}_{v+1})$ and is thus aimed at minimizing the $(v + 1)$-th constrained cost $J_{v+1}$.

Property 2 is a straightforward consequence of Property 1.

*Property* 2. Under the assumption that the DNNs $Q_v$ are exact representations of the state-action value functions $Q_v$, $v = 0, \ldots, C$, by using the control logic (14) a stationary deterministic policy is found, which is lexicographically optimal with respect to the vectorial state-action value function (9).

*Remark* 1. Under the assumptions of Property 2, if the feasible set of the problem (10) is not empty, the lexicographically optimal solution is an optimal solution of the problem (10). Otherwise, i.e., if no solutions exist which satisfy all the constraints, the lexicographic approach computes a sub-optimal policy which is not a feasible solution of (10) but satisfies the maximum number of ordered constraints. In this case, since the algorithm aims at satisfying the constraints according to their priority, the solution generally depends on their ordering.

*Remark* 2. If different thresholds $\bar{K}_c$ are required, there is no need of re-training the DNNs: the desired behavior can be obtained by using the already trained DNNs with the lexicographic action selection according to the new values of $K_c = \bar{K}_c/(1-\gamma)$, $c = 1, \ldots, C$.

### B. Lexicographic Deep Q-Network

As reference algorithms for the algorithm class identified above, we picked the well-known Deep Q-learning with Experience Replay algorithm, also known as Deep Q-Network (DQN) [8], which considers a finite action set. To improve the training process, DQN utilized the replay buffer [18], which stores the state transitions and cost observations occurred at each time-step; the update rule for the DNN is then performed based on the costs contained in the buffer and not on the current observed one.

Fig. 2 presents the lexicographic DQN (L-DQN) algorithm, which accounts for prioritized constraints. As described in Section III.A, the modifications consist in the utilization of additional $C$ DNNs, $Q_c, c = 1, \ldots, C$, to represent the constraint state-action value functions and in the lexicographic action selection. The training phase is the same as in the standard DQN but it is needed for $(C+1)$ DNNs: the primary DNN, minimizing the primary expected total cost, and the constraint DNNs, each one minimizing one of the constraint cost.

As the DNNs are trained, they are ready to be used by the controller. The action selection is performed according to the lexicographic approach. At each time-step $t$, the algorithm of *Function 1* (see Fig. 1) is used to determine the number $v$ of satisfied ordered constraints and the constraint action sets $A_c, c = 1, \ldots, C$. If all the constraints are met, i.e., $v = C$, the action is selected in the set $A_C$ and is aimed at minimizing the primary cost $J_0$; if one or more constraints are not met, i.e., $v < C$, the action is selected in the set $A_v$ and is aimed at minimizing the constraint cost $J_{v+1}$ associated to the first constraint which is not met.

The L-DQN pseudo-code is reported in Fig. 2. As analyzed in [10], DQN, as the original Q-Learning algorithm, tends to overestimate the values of the state-action value function. Even if this problem is not vital in some applications, where obtaining the optimal policy is the main objective, it is of great relevance in the proposed L-DeepRL framework, since it may prevent the algorithm to guarantee the performance requested, in probability, to the controller. The overestimation issue was addressed by the introduction of Double Q-Learning for the tabular algorithm, later translated into Double DQN (D-DQN) for DeepRL solutions [10]. Even if the simulations were run using a lexicographic D-DQN implementation, this section describes the L-DQN algorithm for the sake of readability.

---

*Function 1.* Function for the computation of the discrete constraint action sets in state $s \in S$ observed at time $t$

**Input**: $s, Q_c(\varphi_c(s), u|\boldsymbol{\theta}_c), \forall u \in A_0, c = 0, \ldots, C$ and $K_c, c = 1, \ldots, C$
- Initialize $c = 0$ and $A_v(s) = \emptyset, v = 1, \ldots, C$
- While $c < C$ and $A_{c-1}(s) \neq \emptyset$ do
  - Update $c \leftarrow c + 1$
  - For all $u \in A_{c-1}(s)$ do
    - If $Q_c(\varphi_c(s), u|\boldsymbol{\theta}_c) \leq K_c$, update $A_c(s) \leftarrow A_c(s) \cup \{u\}$
  - If $A_c(s) = \emptyset$ set $v = c - 1$ and $c = C$

**Output**: $v$ and $A_c(s), c = 1, \ldots, v + 1$

Fig. 1. Computation of the discrete constraint action sets.

---

*Algorithm* 1. Lexicographic Deep Q-Network (L-DQN)

*Training*
- Initialize $(C + 1)$ replay buffers $\mathcal{D}_c$ to size $N$, and set minibatch sizes $M_c$ and number of sequences in the minibatches $b = 0$
- Initialize action-value functions $Q_c, c = 0, \ldots, C$, with random weights
- For $c = 0, \ldots, C$
  - For *episode* $= 1, \ldots, M$ do
    - Initialize sequence with random initial state $s_0$ and preprocessed sequences with $\varphi_c(s_0), c = 0, \ldots, C$
    - For time steps $t = 0, \ldots, T$ do
      - With probability $\varepsilon$ select a random action $u_t \in A_c(s_t)$ otherwise select $u_t = \min_{u' \in A_c(s_t)} Q_c(\varphi_c(s_t), u'|\boldsymbol{\theta}_c)$ *(ε-greedy action selection)*
      - Execute action $u_t$ in emulator, observe cost $r_t$ and next state $s_{t+1}$ and set $b = b + 1$
      - Preprocess $\varphi_c(s_{t+1})$
      - Store the transition $\langle \varphi_{c,b}, u_b, r_b, \varphi_{c,b}'\rangle = \langle \varphi_c(s_t), u_t, r_t, \varphi_c(s_{t+1})\rangle$ in $\mathcal{D}_c$
      - Every $\mathcal{T}$ time steps do
        - Sample a minibatch $\mathcal{B}_c$ of $M_c$ random transitions from $\mathcal{D}_c$
        - For each transition $j \in \mathcal{B}_c$
          - Set $y_j = \begin{cases} r_j & \text{for terminal } \varphi_{c,j+1} \\ r_j + \gamma \min_{u \in A_0} Q_c(\varphi_{c,j+1}, u|\boldsymbol{\theta}_c) & \text{otherwise} \end{cases}$
        - Update the critic by minimizing the loss
          $L = \frac{1}{N}\sum_{j \in \mathcal{B}_c}\left(y_j - Q_c(\varphi_{c,j}, u_j|\boldsymbol{\theta}_c)\right)^2$
          *(Training of the c-th critic network — Experience Replay)*

*Lexicographic RL Agent*
- Observe initial state $s_0$
- For $t = 0, \ldots, T$ do
  - Use *Function 1* (see Fig. 1) to compute the number $v$ of met ordered constraints, and the action sets $A_c(s_t), c = 1, \ldots, v + 1$, based on $s_t, Q_c, c = 0, \ldots, C$, and $K_c, c = 1, \ldots, C$ *(Computation of the constraint action sets)*
  - If $v = C$, select
    $u = \min_{u' \in A_C(s_t)} Q_0(\varphi_0(s_t), u'|\boldsymbol{\theta}_0)$
    Otherwise, select
    $u = \min_{u' \in A_v(s_t)} Q_{v+1}(\varphi_{v+1}(s_t), u'|\boldsymbol{\theta}_{v+1})$ *(Lexicographic action selection)*
  - Execute action $u$, observe cost $r_t$ and next state $s_{t+1}$

Fig. 2. Pseudo-code of the L-DQN algorithm.

## III. APPLICATION TO THE CONSTRAINED CART-POLE PROBLEM

The scenario considered to validate the approach consists in the classic cart-pole RL problem, originally presented in [19], that has later become a standard benchmarking environment for RL/DeepRL solutions. The implementation is based on the environment implemented via OpenAI in the Gym toolkit [20], in which the state space is defined by

$$S = \{s = (x\ \dot{x}\ \omega\ \dot{\omega})\ \text{s.t.}\ |x| \leq 2.4m, |\omega| \leq 0.21 rad\}, \quad (15)$$

where $x$ and $\dot{x}$ are the cart position and velocity, respectively, and $\omega$ and $\dot{\omega}$ are the pole angle (with $0\ rad$ defining the straight standing position) and angular velocity, respectively. The two box constraints in (15) define an operative region.

The action space is defined by $A_0 = \{u | u \in \{-10, -5, 0, 5, 10\}\}$, where each action corresponds to applying the specified force, expressed in Newton. A uniform initial distribution $\chi$ was selected in the range $\|s\|_\infty \leq 0.05$. A state is said to be terminal if the cart position or the pole angle are not included in the operative region. In case a terminal state is reached, the cart-pole is re-started in a random position according to the distribution $\chi$.

The primary objective of the lexicographic RL (L-RL) agent consists in maintaining the cart-pole system state within the operative region while minimizing the required force. This objective is captured by the cost function $\rho_0$:

$$\rho_0(s_t, u_t, s_{t+1}) = \begin{cases} |u_t| & \text{if } s_{t+1} \text{ is not terminal} \\ 10 & \text{otherwise} \end{cases}.$$

Regarding the chance-constraints, the one with the highest priority is defined to impose the cart-pole system to maintain the magnitude of the angle $\omega$ within $\pm 0.03 rad$ with a threshold probability $\overline{K}_1$, while the second constraint consists in maintaining the cart position within $\pm 0.1m$ with a threshold probability $\overline{K}_2$. The two cost functions $\rho_1, \rho_2$ penalize the states where the state evolves outside the desired region:

$$\rho_1(s_t) = \begin{cases} 0 & \text{if } |\omega_t| \leq 0.03 \\ 1 & \text{otherwise} \end{cases}, \rho_2(s_t) = \begin{cases} 0 & \text{if } |x_t| \leq 0.1 \\ 1 & \text{otherwise} \end{cases}.$$

As motivated in Section III, the implemented algorithm is the D-DQN, with target DNN trained according to the soft target update method ([21]), with the parameter $\tau$ set to 0.1. All the DNNs were trained with discount factor $\gamma = 0.995$, decaying learning rate $\alpha(t) = 10^{-4} \cdot 0.99^{\max\{1, t-500\}-1}$ and decaying $\varepsilon(t) = 0.5 \cdot 0.99^{\max\{1, t-500\}-1}$. The experience replay was played after every time step, i.e., $\mathcal{T} = 1$. The simulation length was 200 time-steps and the other physical parameters of the cart-pole can be found in [19], [20].

For all the reported tests, a total of 100 episodes with initial state $s_0 \in \chi$ were executed. The left (right) plots of Fig. 3 show the percentage of time that the cart-pole spent in a given position (angle) range. The figures also highlight the desired position and angle ranges $|x| \leq 0.1$ and $|\omega| \leq 0.03$. Table 1 collects the results in terms of percentage of time within the desired position and angle ranges and average absolute value of the force applied during the runs. Figures 3.a)-c) show the results when controlled by only the DNN trained to minimize $J_0$ (minimization of the average used force), $J_1$ (minimization of the angle displacement) and $J_2$ (minimization of the distance from $x = 0$), respectively. All the DNNs are characterized by two hidden layers of 64 neurons with *relu* activation functions, save for $Q_0$ that has 16 neurons on the second layer, and a linear dense output layer. The training required approximatively 400 episodes for each DNN.

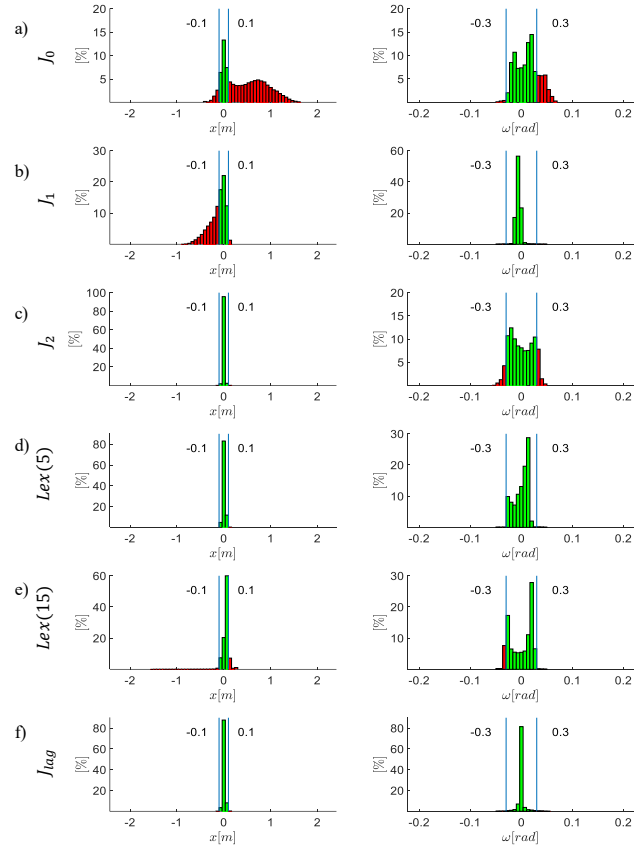

Fig. 3. Percentage of time within various position (left plots) and angle ranges (right plots) with different RL and L-RL agents.

TABLE 1
SIMULATION RESULTS

| Cost function | % of time outside desired positions | % of time outside desired angles | Average applied force |
|---|---|---|---|
| $J_0$ | 73.8% | 22.3% | $0.29N$ |
| $J_1$ | 48.1% | 0.7% | $0.39N$ |
| $J_2$ | 0.2% | 15.6% | $2.97N$ |
| $Lex(5)$ | 0.3% | 0.6% | $1.39N$ |
| $Lex(15)$ | 12.2% | 8.5% | $1.17N$ |
| $J_{lag}$ | 0.5% | 1.3% | $2.54N$ |

Fig. 3.a) shows that the control policy found by minimizing $J_0$ is such that the cart position and angle are often on the positive $x$ and $\omega$ values, leading to a percentage of time spent outside the desired region of 72.8% for the position range and 12.3% for the angle, as reported in Table 1, with spent average force of $0.29N$. Fig. 3.b) shows that, under the cost $J_1$, the angle is almost never outside the desired angle region (0.7% of the time-steps), the percentage of time spent outside the desired position region is 48.1% and the spent average force is $0.39N$.

As shown in Fig. 3.c), under the cost $J_2$ the controller limits the time outside the position range to 0.2% at the price of a larger effort, $2.97N$. The angle lies outside the desired region 15.6% of the time.

Figures 3.d)-e) show the results with L-RL agents, with thresholds $\bar{K}_1 = \bar{K}_2 = 0.05$ and $\bar{K}_1 = \bar{K}_2 = 0.15$, denoted with $Lex(5)$ and $Lex(15)$, respectively. The L-RL agents exploit the same 3 DNNs trained for the previous tests and, in each state, use one of the DNNs to minimize the corresponding cost. Fig. 3.d) shows that, with the first L-RL agent, the cart-pole is almost never outside the desired region (less than 1% for both position and angle) by spending an average force of $1.39N$, significantly smaller than the one spent under $J_2$ as the L-RL agent uses also the DNN trained for the force minimization objective. Fig. 3.e) shows that also with the second L-RL agent the cart-pole is outside the desired region for less than its prescribed percentage of time (12.2% for the angle, 8.5% for the position). As the prescribed percentages are smaller for the latter L-RL agent, the average spent force is reduced to $1.17N$.

During the episodes, the first L-RL agent, $Lex(5)$, used $Q_0$ (trained based on the primary cost $\rho_0$, i.e., to minimize the control effort) to select the control action in 11% of the time-steps, $Q_1$ (trained based on the angle cost $\rho_1$) in 4% and $Q_2$ (trained based on the position cost $\rho_2$), in 85%. The second L-RL agent, $Lex(15)$, which has lower probability thresholds, manages to increase the percentage of time in which $\rho_0$ is minimized: it uses $Q_0$, $Q_1$ and $Q_2$ in 27%, 3% and 70% of the time-steps, respectively.

For comparison purposes, Fig. 3.f) shows the results with a RL agent aimed at minimizing the multi-objective cost function $J_{lag} := \lambda [J_0 \; J_1 \; J_2]^T$, where $\lambda = [1 \; 5 \; 25]$ is the vector of the Lagrangian weights associated to the cost functions $J_i$'s. To achieve the prescribed percentages of 5%, the weights were tuned by extensive grid-search during the training phase of a DNN (analogous to the ones trained for the previous simulations), which required approximately 600 episodes. By using this DNN, the RL agent manages to achieve similar performance with respect to the L-RL agent with the same targets ($Lex(5)$) at the price of a larger control effort, equal to $2.54N$. Better results can be obtained with finer weight tuning techniques, which are out of the scope of the paper. Conversely, it is important to remark that the DNN should be trained again to aim at the prescribed percentages of 15% – and at a consequently lower control effort.

## IV. Conclusions and Future Works

This paper proposed an extension of the lexicographic approach to the DeepRL framework, showing how it can be used to design chance-constrained controllers. The main advantages with respect to standard methods are i) that no additional tuning of hyper-parameters is required in the training phase to cope with the constraints and ii) that the probability with which the constraints are met can be changed without the need of re-training the DNNs.

Future work is aimed at extending the lexicographic approach to online solutions and continuous action space scenarios by extending actor-critic methods [21].